\documentclass[11pt]{article}
\usepackage{geometry}
\usepackage{coling2020}
\usepackage{times}
\usepackage{url}
\usepackage{latexsym}
\usepackage{microtype}
\usepackage{tabularx}
\usepackage{booktabs}

 \usepackage[normalem]{ulem}
 \useunder{\uline}{\ul}{}
 \usepackage{wrapfig}

\usepackage{amssymb}
\usepackage{amsmath}
\usepackage[ruled,vlined]{algorithm2e}
\usepackage[labelfont=bf,textfont=md]{caption}
\usepackage{multirow}
\usepackage{pgfplots}
\pgfplotsset{width=11cm,compat=1.9}
\usepackage{todonotes}

\colingfinalcopy 

\title{GM-CTSC at SemEval-2020 Task 1: Gaussian Mixtures Cross Temporal Similarity Clustering}

\author{Pierluigi Cassotti \\
  University of Bari, Italy \\
  {\tt pierluigi.cassotti@uniba.it} \\\And
  Annalina Caputo \\
  Dublin City University, Ireland \\
  {\tt annalina.caputo@dcu.ie} \\\AND
  Marco Polignano \\
  University of Bari, Italy \\
  {\tt marco.polignano@uniba.it} \\\And
  Pierpaolo Basile \\
  University of Bari, Italy \\
  {\tt pierpaolo.basile@uniba.it} \\}

\date{}

\begin{document}
\maketitle
\begin{abstract}
This paper describes the system proposed for the SemEval-2020 Task 1: Unsupervised Lexical Semantic Change Detection. 
We focused our approach on the detection problem. Given the semantics of words captured by temporal word embeddings in different time periods, we investigate the use of unsupervised methods to detect when the target word has gained or loosed senses.
To this end, we defined a new algorithm based on Gaussian Mixture Models to cluster the target similarities computed over the two periods. We compared the proposed approach with a number of similarity-based thresholds. We found that, although the performance of the detection methods varies across the word embedding algorithms, the combination of Gaussian Mixture with Temporal Referencing resulted in our best system.

\end{abstract}

\section{Introduction}
\label{intro}
The recent development in word embeddings, and their increasing capability to capture lexical semantics
has inspired the application of these methods to new tasks and introduced new challenges. 
The diachronic analysis of language is one of such linguistic tasks that has benefited from the advantages of these new methods, i.e. the capability to build semantic representations of words by skimming through large corpora spanning multiple time periods.
SemEval 2020 Task 1 \cite{schlechtweg2020semeval} addresses the current lack of a systematic approach for the evaluation of automatic methods for the diachronic analysis by proposing a common evaluation framework that comprises two tasks and covers four different languages (German, English, Latin, and Swedish).
Given two corpora $C_1$ and $C_2$ for two periods $t_1$ and $t_2$, Subtask 1 requires participants to classify a set of target words in two categories: words that have lost or gained senses from $t_1$ to $t_2$ and words that did not, while Subtask 2 requires participants to rank the target words according to their degree of lexical semantic change between the two periods.
We tackle the problem of automatically detecting lexical semantic changes with approaches that rely on temporal word embeddings.
These approaches create a word vector representation for each time period by exploiting a shared semantic space. Similarity measures can then be used to capture the extent of a word semantic change between two time lapses.
Some temporal word embedding techniques adopt a two-step approach, where they first learn separate word embeddings for each time period and then align the word vectors across multiple time periods \cite{Hamilton2016}. Other \textit{dynamic} approaches incorporate the alignment directly into the learning stage via the optimisation function \cite{Tahmasebi2018}.
Dynamic word embeddings can be further categorised according to the constraint imposed on the alignment. 
The \textit{explicit} alignment adopts a conservative approach to the semantic drift that a word can undergo by posing a limit to the distance between the word vectors belonging to the two temporal spaces.
In the \textit{implicit} alignment, there is no need for explicit constraint since the alignment is automatically performed by sharing the same word context vectors across all the time periods.

In this work, we focus on dynamic word embeddings by exploring methods based on both explicit, such as Dynamic Word2Vec \cite{Yao2018}, and implicit alignment, namely Temporal Random Indexing \cite{Basile} and Temporal Referencing \cite{Dubossarsky2019}.
We analyse the use of different similarity measures to determine the extent of a word semantic change and
compare the cosine similarity with Pearson Correlation and the neighborhood similarity \cite{Shoemark2019}.
While these similarity measures can be directly employed to generate a ranked list of words for Subtask 2, their adoption in Subtask 1 requires further manipulation.
We introduce a new method to classify changing vs. stable words by clustering the target similarity distributions via Gaussian Mixture Models.
We describe the embedding models and the clustering algorithm in Section 2, while Section 3 provides details about the hyper-parameter selection. Section 4 reports the results of the task evaluation followed by some concluding remarks in Section 5.

\section{GM-CTSC}\label{Section2}
We model the problem of automatic detection of semantic change by exploiting temporal word embeddings $E_{i}: w \rightarrow \mathbb{R}^d$ that project each word $w$ in the vocabulary $V$ into a $d$-dimensional semantic space. Given two different time periods $t_1$ and $t_2$, we create two embeddings $E_1$ and $E_2$.
We investigate several models to compute temporal words embeddings:
\begin{description}
  \item \textbf{Dynamic Word2Vec (DW2V)} \cite{Yao2018} simultaneously learns time-aware embeddings by aligning and reducing the dimensionality of time-binned Positive Point-wise Mutual Information matrices. 
  \item \textbf{Temporal Random Indexing (TRI)} \cite{Basile} implicitly aligns co-occurrence matrices by using the same random projection for all the temporal bins.
  \item \textbf{Collocations} extracts for each word and each time period the set of relevant collocations through Dice score. 
  As similarity function, we measure the cosine similarity between the sets of collocations belonging to the two different time periods. More details are reported in \newcite{BasileKronos}.
  \item \textbf{Temporal Referencing (TR)} \cite{Dubossarsky2019} used only in the post-evaluation, it consists in a modified version of Word2Vec Skipgram that adds a temporal 
referencing to target vectors, keeping context vectors unchanged.
\end{description}

A similarity measure between vectors in the two temporal spaces is adopted to compute the extent of the semantic drift of the target words. We explored several similarity measures:

\begin{description}
  \item {\textbf{Cosine similarity (CS)}} is the cosine of the angle between two vectors.
  \item {\textbf{Pearson correlation (PC)}} measures the linear correlation between two variables, in case of centred vectors (with zero means) is equivalent to the cosine similarity.
  \item {\textbf{Neighborhood similarity (NS)}} computes two $k$-neighbour sets $nbrs_k(E_1(w))$ and $nbrs_k(E_2(w))$ and the union set \( \mathcal{U} = nbrs_k(E_1(w)) \cup nbrs_k(E_2(w))\). Two second-order vectors, one for each word representation $u_j$, are created. The components of $u_i$ are the cosine similarity between the vector $v_j$\footnote{Where $v_j$ is the vector representation for the word generated by $E_j$ and $j$ is the time period.} and the i-th element of $\mathcal{U}$: $u_{j_i}=cos(v_j, \mathcal{U}(i))$. The Neighborhood similarity is the cosine similarity between the second-order vectors. In all the experiments we set $k=25$.
\end{description}

\subsection{Subtask 2}
In Subtask 2, we use one of the three similarity measures ($CS$, $PC$, $NS$) to compute the set of target similarities $\mathcal{S}=\{sim(E_1(w),E_2(w)) \mid w \in T \}$. Then, we rank the target words according to the distance, computed as: $1 - \mid sim(E_1(w), E_2(w))\mid$.

\subsection{Subtask 1: Gaussian Mixture Clustering} 
Subtask 1 requires a further step: given $\mathcal{S}$, the set of target similarities, we need to predict the target labels. 
The aim is to assign either of the two classes, 0 (stable) or 1 (change), to each target word of a given language.
Once we compute the set of target similarities $\mathcal{S}$, we want to find a way to assign the corresponding label. 
We assume that low similarities suggest changing words and high similarities indicate stable words. 

Gaussian Mixture Models (GMMs) allow to build probabilistic models for representing the Gaussian distribution of stable and changed targets.
We use GMMs\footnote{\url{https://scikit-learn.org/stable/modules/generated/sklearn.mixture.GaussianMixture.html}} to model the  density  of the distributions of the similarities of targets as a weighted sum of two Gaussian densities \cite{huang2017model}:

\begin{equation}
\label{eq:gmm}
f(\mathcal{S}) = \sum_{m=0}^{M}\pi_m\phi(\mathcal{S}|\mu_m,\Sigma_m)
\end{equation}
where $M$ is the number of mixture components, $\phi(\mathcal{S}|\mu_m,\Sigma_m)$ is the Gaussian density with mean vector $\mu_m$ and covariance matrix $\Sigma_m$, and $\pi_m$ is the prior probability for the $m$-th component.
Additional constraints can be applied to the covariance matrix in Eq. \ref{eq:gmm}. 
In our experiments, we allow each component to have its own covariance matrix. 

For our purpose, we speculate that the distribution of target similarities is a mixture of two densities, i.e. representing the stable and changing words. Consequently, we fixed the number of the mixture components in the GMMs to two.
We initially randomly assign a label (stable/changing) to each density distribution.
Let $\mu_0$ and $\mu_1$ be the means of the two Gaussians associated with the ``stable'' and ``changing'' labels respectively. If $\mu_0 < \mu_1$ (i.e. the similarity mean of the distribution labelled as ``stable'' is lower than the mean of distribution labelled as ``changing''), we invert the labels.
Alg. \ref{algorithm:algorithm1} can be used for properly label each word of the target vocabulary.

\begin{algorithm}[h!]

\SetAlgoLined
\SetKwInOut{Input}{input}
\SetKwInOut{Output}{output}
\Input{$\mathcal{S}$}
\Output{labels}
 $\mathcal{N}(\mu_0,\sigma_0),\mathcal{N}(\mu_1,\sigma_1),labels \longleftarrow GaussianMixtures(\mathcal{S})$\;
 \If{$\mu_0 < \mu_1$}{
    $labels \longleftarrow 1-labels$\;
 }
 \caption{Assign labels}
 \label{algorithm:algorithm1}
\end{algorithm}

In order to set the best parameters for each language and model, we rely on the GMMs log likelihood, which is generally used for estimating the clusters quality:
\begin{equation}
 \ell(\theta \mid \mathcal{S}) = log \sum_{m=0}^{M}\pi_m\phi(\mathcal{S}\mid \mu_m, \Sigma_m)   
\end{equation}

where $\theta$ are the parameters of the GMM.
For each language, we select the best model configuration to submit at the challenge using the GMMs log likelihood $\ell(\theta \mid \mathcal{S})$. 
We improperly use this approach for choosing parameters across different models (different sets of similarities $\mathcal{S}$), as we do not have validation set for tuning the parameters. 
We will investigate this limitation as future work.
The selected models and hyper-parameters are reported in Tab. \ref{table:config}. In particular, we use cosine similarity, Pearson correlation and Neighborhood similarity for computing the targets similarities in  $Overall_{CS}$, $Overall_{PC}$ and $Overall_{NS}$ runs, respectively. In $DW2V$ and $TRI$ runs we use always cosine similarity.

\section{Experimental Setup}
In all the runs, we do not pre-process data and we use a context window size of 5 while analyzing sentences. The $TR$ model\footnote{We add this model during the post-evaluation.} has been adopted into its original implementation\footnote{\url{https://github.com/Garrafao/TemporalReferencing}}, as the $TRI$\footnote{\url{https://github.com/pippokill/tri}} approach and $DW2V$\footnote{\url{https://github.com/yifan0sun/DynamicWord2Vec}} one.
For runs involving $TRI$, we experimented with a varying vector size from $200$ to $1,000$.
Moreover, we investigated (1) the initialization of the count matrix at time $j$ with the matrix at time $j-1$, (2) the contribution of positive-only projections, and (3) the application of PPMI weights, as explained in \newcite{QasemiZadeh2016}.
For $DW2V$, we use the parameter setting proposed in \newcite{Yao2018}. We set \(\lambda=10\), \(\tau=50\), \(\gamma=100\), \(\rho=50\) and experimented with a number of iterations from one to five.
As vocabulary, we kept the top 50,000 most frequent tokens for both $TRI$ and $DW2V$. 
In the $TR$ runs, we set the vector size to $100$, and we experimented eight iterations for English and Latin, and four for German and Swedish. We use $20$ negative samples, keeping only the tokens that occur at least $10$ times.
All the other parameters used for configuring the models are reported in Tab. \ref{table:config}.

\begin{center}
\resizebox{300px}{!}{
 \begin{tabular}{c c c c c c} 
\toprule
 Run & Configuration & English & German & Latin & Swedish\\
 \midrule
 \multirow{2}{*}{$Overall_{CS}$} & Model & DW2V & Collocation  & DW2V  & DW2V \\
 & Parameters & it=3 & - & it=3 & it=4\\
 \midrule
 
 \multirow{2}{*}{$Overall_{PC}$} & Model & DW2V & DW2V  & DW2V  & DW2V \\
 & Parameters & it=3 & it=4 & it=3 & it=4\\
 \midrule
 \multirow{2}{*}{$Overall_{NS}$}  & Model & DW2V & DW2V  & DW2V  & DW2V \\
 & Parameters & it=3 & it=1 & it=3 & it=4\\
 \midrule
 \multirow{2}{*}{$TRI$} & \multirow{2}{*}{Parameters} & k= 400 & k=1000 & k=1000 & k=1000\\
 & & pw=False & pw=True & pw=True & pw=True\\
 \midrule
 $DW2V$ & Parameters & it=3 & it=4 & it=3 & it=4\\
 \bottomrule
\end{tabular}
}

\captionof{table}{Hyper-parameters and models selected for each run. \textit{it} is the number of iterations, \textit{k} is the embedding size, \textit{pw} the use of PPMI weights}
\label{table:config}

\end{center}

\section{Results}
Tab. \ref{table:res} reports the main results obtained by the different models. 
It shows the results obtained from the official submissions at the challenge and the results obtained by the $TR$ approach performed during the post-evaluation phase. The results obtained for the Subtask 1 are reported using the accuracy metric, while for the Subtask 2, the Spearman's rank-order correlation coefficients are used.

Considering the results of the evaluation phase, the models show not consistent behaviors. 
$TRI$ showed the best performance when considering ``all the languages'' for both Subtasks, although in Subtask 1 it is not able to overcome \textit{Baseline2}. 
Focusing on Subtask 1, if we consider each language in isolation, we see that $DW2V$ gives the best results for English\footnote{Please, note that for EN, LA and SW $Overall_{CS}$ and $DW2V$ coincide} while $Overall_{PC}$ (Collocation with cosine similarity) is our best system for German language, although it is not able to overcome \textit{Baseline2}. 
$TRI$ is the best system for Latin, although outperformed by \textit{Baseline1}, and Sweden languages.
In Subtask 2, the best English score was reported by $Overall_{NS}$.
Simlarly to Subtask 1, $Overall_{CS}$ performed the best in German language.
For Latin and Sweden, $TRI$ provided the best results, and interestingly, it is one of the few systems that did not generate a negative correlation.
For Sweden language in particular, it is interesting to notice that $TRI$ generated the best result among all the task participants.

At the end of the challenge, when the labelled test set was released, we performed more experiments reported in the \textit{post-evaluation} row.
In this phase we run an additional system, $TR$, which outperformed all the previous reported approaches, including both baselines.
The only exception is for Latin, in which for Subtask 1 $Baseline1$ achieves $0.650$ accuracy in comparison to $0.525$ of $TR$. 
Comparing $TR$ and $TRI$, which are both based on implicit alignment, the former is a prediction-based model while the is latter a count-based one. 
Moreover, $TR$ creates a temporal word embedding only for the target words rather than for the whole vocabulary. Consequently, this results in better word embeddings for all the words in the vocabulary that do not have a temporal reference. 
These differences allow $TR$ to achieve better results than the other models.

\begin{table}[]
\centering
\resizebox{430px}{!}{
\begin{tabular}{l l l l l l c l l l l l }
\toprule
\multicolumn{1}{c}{} & \multicolumn{5}{c}{\textbf{Subtask 1}} & & \multicolumn{5}{c}{\textbf{Subtask 2}} \\ 
\cmidrule(lr){2-6} \cmidrule(lr){8-12}
\multicolumn{1}{c}{\textbf{System}} & \multicolumn{1}{c}{\textbf{\begin{tabular}[c]{@{}c@{}}All\\ Lang.\end{tabular}}} & \multicolumn{1}{c}{\textbf{EN}} & \multicolumn{1}{c}{\textbf{GE}} & \multicolumn{1}{c}{\textbf{LA}} & \multicolumn{1}{c}{\textbf{SW}} & & \multicolumn{1}{c}{\textbf{\begin{tabular}[c]{@{}c@{}}All\\ Lang.\end{tabular}}} & \multicolumn{1}{c}{\textbf{EN}} & \multicolumn{1}{c}{\textbf{GE}} & \multicolumn{1}{c}{\textbf{LA}} & \multicolumn{1}{c}{\textbf{SW}} \\ 
\midrule
\multicolumn{1}{l}{\textit{$Baseline1$}} & 0.439 & 0.432 & 0.417 & \textit{\textbf{0.650}} & 0.258 & & -0.083 & -0.217 & 0.014 & 0.020 & -0.150 \\ 
\multicolumn{1}{l}{\textit{$Baseline2$}} & \textit{0.613} & 0.595 & \textit{0.688} & 0.525 & 0.645 & & 0.144 & 0.022 & 0.216 & \textit{0.359} & -0.022 \\ 
\midrule
\multicolumn{1}{l}{\textit{$Overall_{CS}$}} & 0.509 & \textit{0.622} & 0.500 & 0.400 & 0.516 & & 0.111 & 0.252 & \textit{0.415} & -0.183 & 0.041 \\ 
\multicolumn{1}{l}{\textit{$Overall_{PC}$}} & 0.533 & 0.595 & 0.646 & 0.375 & 0.516 & & 0.056 & 0.272 & 0.168 & -0.135 & -0.080 \\ 
\multicolumn{1}{l}{\textit{$Overall_{NS}$}} & 0.508 & 0.568 & 0.542 & 0.375 & 0.548 & & 0.035 & \textit{0.298} & -0.059 & -0.179 & 0.078 \\ 
\multicolumn{1}{l}{\textit{$Collocation$}} & 0.513 & 0.486 & 0.500 & 0.550 & 0.516 & & 0.273 & 0.144 & \textit{0.415} & 0.194 & 0.340 \\ 
\multicolumn{1}{l}{\textit{$DW2V$}} & 0.541 & \textit{0.622} & 0.625 & 0.400 & 0.516 & & 0.098 & 0.252 & 0.366 & -0.183 & -0.041 \\ 
\multicolumn{1}{l}{\textit{$TRI$}} & 0.554 & 0.486 & 0.479 & 0.475 & \textit{0.774} & & 0.296 & 0.211 & 0.337 & 0.253 & \textit{0.385} \\ 
\midrule
\multicolumn{1}{l}{\textit{\begin{tabular}[c]{@{}l@{}}$TR$\\ (post-eval.)\end{tabular}}} & \textbf{0.704} & \textbf{0.703} & \textbf{0.812} & 0.525 & \textbf{0.774} & & \textbf{0.496} & \textbf{0.304} & \textbf{0.722} & \textbf{0.395} & \textbf{0.562} \\ 
\bottomrule
\end{tabular}
}
\captionof{table}{
Results obtained by our models during the official competition and during the post-evaluation phase.
For the Subtask 1 the results represent the accuracy score. Spearman's rank-order correlation coefficients are used for the Subtask 2.}
\label{table:res}
\end{table}
During the post-evaluation we decided to investigate also the role of GMMs for class labeling (Sec. \ref{Section2}).
We compared GMMs with semi-manual thresholds \(\mu_{\mathcal{S}}\), \(\mu_{\mathcal{S}}-\sigma_{\mathcal{S}}\),
\(\mu_{\mathcal{S}}+\sigma_{\mathcal{S}}\) and
Winsorizing \cite{kokic1994optimal} computing \(\mu_{\boldsymbol{S}}\) and \(\sigma_{\boldsymbol{S}}\) on data provided for Subtask 1, where $\mu_\mathcal{S}$, $\sigma_\mathcal{S}$ are the mean and the standard deviation computed on the similarity set $\mathcal{S}$.
Figure \ref{figure:figure1} reports the different accuracy scores obtained by the five methods for the $TRI$, $Collocation$, $DW2V$, $TR$ approaches.
The scores for the GMMs strategy are close to those obtained by
\(\mu_{\mathcal{S}}\) for TRI and Collocation.
While GMMs outperforms \(\mu_{\mathcal{S}}+\sigma_{\mathcal{S}}\) in every run,
 \(\mu_{\mathcal{S}}-\sigma_{\mathcal{S}}\) seems to work better than GMMs except that in $TR$.
Winsorizing work better in $TRI$ and $Collocation$ than GMMs. GMMs outperforms Winsorizing in $DW2V$ and $TR$.
 These results are not clear enough to advocate for a specific threshold.
Consequently, further analysis will be part of future work in order to understand what is the better threshold that could be included in the GMMs process.

\begin{center}
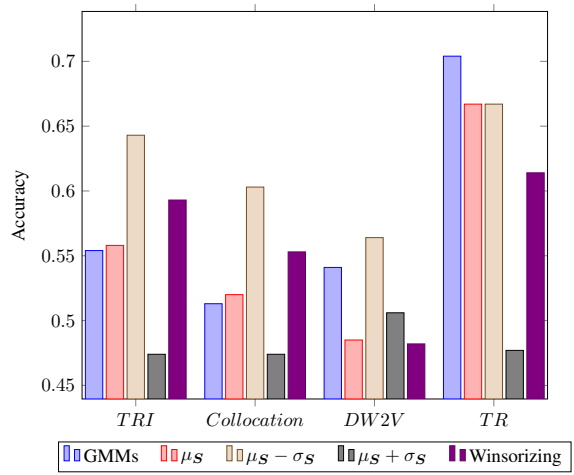

\scalebox{0.65}{
\begin{tikzpicture}
\begin{axis}[
    symbolic x coords={$TRI$,$Collocation$,$DW2V$,$TR$},
    xtick=data,
	ylabel=Accuracy,
	enlargelimits=0.15,
	legend style={at={(0.5,-0.11)},
	anchor=north,legend columns=-1},
	ybar
]
\addplot coordinates {($TRI$,0.554) ($Collocation$,0.513) ($DW2V$,0.541) ($TR$,0.704)};
\addplot coordinates {($TRI$,0.558) ($Collocation$,0.520) ($DW2V$,0.485) ($TR$,0.667)};
\addplot coordinates {($TRI$,0.643) ($Collocation$,0.603) ($DW2V$,0.564) ($TR$,0.667)};
\addplot coordinates {($TRI$,0.474) ($Collocation$,0.474) ($DW2V$,0.506) ($TR$,0.477)};
\addplot coordinates {($TRI$,0.593) ($Collocation$,0.553) ($DW2V$,0.482) ($TR$,0.614)};
\legend{GMMs$\quad$,\(\mu_{\boldsymbol{S}}\)$\quad$, \(\mu_{\boldsymbol{S}}-\sigma_{\boldsymbol{S}}\)$\quad$, \(\mu_{\boldsymbol{S}}+\sigma_{\boldsymbol{S}}\)$\quad$,Winsorizing}
\end{axis}
\end{tikzpicture}
}
\captionof{figure}{Accuracy scores in Subtask 1 using different class labeling strategies: GMMs, \(\mu_{\boldsymbol{S}}\), \(\mu_{\boldsymbol{S}}-\sigma_{\boldsymbol{S}}\), \(\mu_{\boldsymbol{S}}+\sigma_{\boldsymbol{S}}\) and Winsorizing using mean and standard deviation.}
\label{figure:figure1}

\end{center}

\section{Conclusions}
We described the runs we submitted to the SemEval-2020 Task 1: Unsupervised Lexical Semantic Change Detection. 
This paper has two main contributions. 
We reported a comparison of some of the most recent approaches to model lexical semantic change with temporal word embeddings, and we experimented with an automatic unsupervised procedure to classify changing and stable words. Results show that implicit alignment works generally better in modelling the lexical semantic change. 
In future works we plan to carry out an analysis on unlemmatised corpora and gauge a better understanding of the impact of Gaussian Mixture Clustering for unsupervised lexical semantic change detection.

\bibliographystyle{coling}
\bibliography{semeval2020}

\end{document}